\documentclass[10pt,twocolumn,letterpaper]{article}

\usepackage{wacv}
\makeatletter
\@namedef{ver@everyshi.sty}{}
\makeatother
\usepackage{times}
\usepackage{epsfig}
\usepackage{graphicx}
\usepackage{amsmath}
\usepackage{amssymb}

\usepackage[utf8]{inputenc}

\usepackage[inline]{enumitem}
\usepackage{array}
\usepackage{tikz}
\usepackage{pgfplots}

\usepackage{subcaption}

\def\thicknesslineplot{thick}

\def\mythickness{ultra thick}

\def\mythicknesssmall{thick}
\def\thicknesslineplotThree{very thick}
\def\boxplotheight{8.5cm}
\def\boxplotwidth{7.5cm}
\def\boxplotheightreal{3.5cm}

\definecolor{gtcolor}{rgb}{0,0,0}
\definecolor{ourcolor}{rgb}{0,0.4,0.6}
\definecolor{fitzgibboncolor}{rgb}{0,0.7,0.7}
\definecolor{dingcolor}{rgb}{1,0.2,0.3}
\definecolor{kukelovacolor}{rgb}{0,0.6,0}
\definecolor{bougnouxcolor}{rgb}{0,0.7,0.7}
\definecolor{valtonenoernhagcolor}{rgb}{0.7,0,0.7}
\definecolor{inliercolor}{rgb}{0.2,1,0.1}
\definecolor{outliercolor}{rgb}{1,0,0}
\definecolor{smallcolor1}{rgb}{0.,0,0}
\definecolor{smallcolor2}{rgb}{0,0,0}
\definecolor{hiddencolor}{rgb}{0.1,0.9,0.1}
\definecolor{elimcolor}{rgb}{0.6,0.1,0.6}

\usepackage{standalone}
\input{tikz-pgfplots.settings}
\pgfplotsset{
    table/search path={graphs},
}

\usepackage{bm}
\renewcommand{\vec}[1]{\bm{#1}}
\providecommand{\mat}[1]{\bm{#1}}

\usepackage{mathtools}

\newcommand{\te}[1]{\text{#1}}
\DeclarePairedDelimiterX{\norm}[1]{\lVert}{\rVert}{#1}
\DeclarePairedDelimiterX{\abs}[1]{\lvert}{\rvert}{#1}
\newcommand{\fr}[2]{\frac{#1}{#2}}

\newcommand{\T}{T}
\DeclareMathOperator{\tr}{tr}

\DeclareMathOperator{\diag}{diag}
\makeatletter
\newcommand{\HUGE}{\bBigg@{3}}
\newcommand{\vast}{\bBigg@{4}}
\newcommand{\Vast}{\bBigg@{5}}
\makeatother

\newcommand{\Hy}{\ensuremath{\mat{H}_{y}}}
\graphicspath{{images/}}

\def\wacvPaperID{699} 

\wacvfinalcopy 

\ifwacvfinal
\def\assignedStartPage{1} 
\fi

\ifwacvfinal
\usepackage[breaklinks=true,bookmarks=false]{hyperref}
\else
\usepackage[pagebackref=true,breaklinks=true,colorlinks,bookmarks=false]{hyperref}
\fi
\usepackage[capitalize,noabbrev]{cleveref}

\ifwacvfinal
\else
\fi

\begin{document}

\title{Efficient Real-Time Radial Distortion Correction for UAVs}

\author{%
Marcus Valtonen \"{O}rnhag${}^1$,
Patrik Persson${}^1$,
M{\aa}rten Wadenb\"{a}ck${}^2$,
Kalle {\AA}str{\"o}m${}^1$,
Anders Heyden${}^1$\\[0.2cm]
\begin{minipage}[c]{0.4\textwidth}
    \centering
    ${}^1$Centre for Mathematical Sciences\\
    Lund University
\end{minipage}
\begin{minipage}[c]{0.4\textwidth}
    \centering
    ${}^2$Department of Electrical Engineering\\
    Link\"{o}ping University
\end{minipage}
	\\[0.4cm]
{\tt\small marcus.valtonen\_ornhag@math.lth.se}
}

\makeatletter
\let\thetitle\@title
\let\theauthor\@author
\makeatother

\maketitle

\begin{abstract}
In this paper we present a novel algorithm for onboard radial distortion correction for unmanned
aerial vehicles (UAVs) equipped with an inertial measurement unit (IMU), that runs in real-time.
This approach makes calibration procedures redundant, thus allowing for exchange of optics
extemporaneously.
By utilizing the IMU data, the cameras can be aligned with the gravity direction. This allows
us to work with fewer degrees of freedom, and opens up for further intrinsic calibration.
We propose a fast and robust minimal solver for simultaneously estimating the focal length,
radial distortion profile and motion parameters from homographies. The proposed solver is tested
on both synthetic and real data, and perform better or on par with state-of-the-art methods
relying on pre-calibration procedures.
\end{abstract}

\section{Introduction}
In epipolar geometry, the relative pose of two uncalibrated camera views is encoded algebraically as the fundamental matrix~$\mat{F}$ concomitant with the two views.
When trying to estimate $\mat{F}$ from point correspondences, it is well-known that the minimal case---\ie{} the smallest number of point correspondences for which there exists at most finitely many solutions---uses seven point correspondences~\cite{hartley-zisserman}.
By using eight point correspondences instead of seven, the estimation problem results in a system of eight linear equations, which can be solved fast and in a numerically robust manner using the singular value decomposition (SVD)~\cite{hartley-1997-pami}.
To solve the minimal case, \ie{} using only seven point correspondences, one must conjoin the seven linear equations with one cubic equation emanating from the rank constraint~$\det{\mat{F}} = 0$.
In the case of calibrated cameras, the minimal case involves only five point correspondences;
however, the corresponding system of polynomial equations now contains ten cubic equations, and the
complexity of the solver increases further~\cite{nister-pami-2004}.

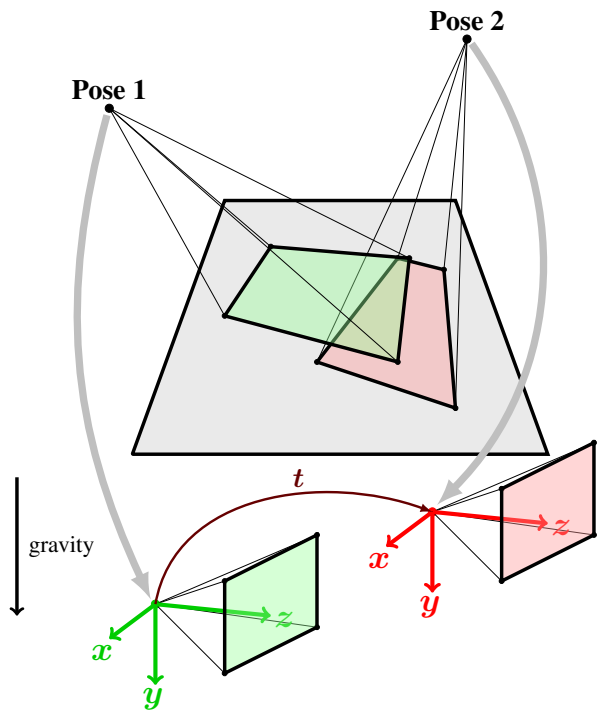
\begin{figure}[h!]
\centering
\begin{tikzpicture}
  \filldraw (-4,-2.5) coordinate (f1) circle (0.5pt);
  \filldraw (-2,3) coordinate (f2) circle (0.5pt);
  \filldraw (3,3) coordinate (f3) circle (0.5pt);
  \filldraw (5,-2.5) coordinate (f4) circle (0.5pt);
  \filldraw [line width = \linethicknessframe, fill=gray!40, fill opacity=0.4] (f1) -- (f2) -- (f3) -- (f4) -- (f1);

  \filldraw (0,-0.5) coordinate (b2a) circle (1.5pt);
  \filldraw (1.75,1.75) coordinate (b2b) circle (1.5pt);
  \filldraw (2.75,1.5) coordinate (b2c) circle (1.5pt);
  \filldraw (3,-1.5) coordinate (b2d) circle (1.5pt);
  \filldraw (3.25, 6.5) coordinate (a2) circle (2.5pt) node[yshift=4mm] {\LARGE \textbf{Pose 2}};

  \filldraw [line width = \linethicknessframe, fill=red!40, fill opacity=0.4] (b2a) -- (b2b) -- (b2c) -- (b2d) -- (b2a);
  \draw [line width=\linethickness] (a2) -- (b2a);
  \draw [line width=\linethickness] (a2) -- (b2b);
  \draw [line width=\linethickness] (a2) -- (b2c);
  \draw [line width=\linethickness] (a2) -- (b2d);

  \filldraw (-2,0.5) coordinate (b3a) circle (1.5pt);
  \filldraw (-1,2) coordinate (b3b) circle (1.5pt);
  \filldraw (2,1.75) coordinate (b3c) circle (1.5pt);
  \filldraw (1.75,-0.5) coordinate (b3d) circle (1.5pt);
  \filldraw (-4.5, 5) coordinate (a3) circle (2.5pt) node[yshift=4mm] {\LARGE \textbf{Pose 1}};

  \filldraw [line width = \linethicknessframe, fill=green!40, fill opacity=0.4] (b3a) -- (b3b) -- (b3c) -- (b3d) -- (b3a);
  \draw [line width=\linethickness] (a3) -- (b3a);
  \draw [line width=\linethickness] (a3) -- (b3b);
  \draw [line width=\linethickness] (a3) -- (b3c);
  \draw [line width=\linethickness] (a3) -- (b3d);

  \draw[->, line width=\linethicknessframe] (-6.5,-3) -- (-6.5,-6) node [midway, right,xshift=1mm] {\Large gravity};
  \filldraw[white] (-6.7,-5) coordinate (dummy) circle (0pt);  

  \coordinate[yshift=-1.5mm,xshift=-0.5mm] (specRoot) at (a3);
  \coordinate (testTreeRoot) at (-3.6,-5.65);
  \draw[-latex, gray!50!white, line width=1ex]  (specRoot) to[in=115,out=255] (testTreeRoot);

  \coordinate[yshift=-1mm,xshift=1mm] (testTreeRoot2) at (a2);
  \coordinate (specRoot2) at (2.6,-3.6);
  \draw[latex-, gray!50!white, line width=1ex]  (specRoot2) to[in=-55,out=45] (testTreeRoot2);

  \begin{scope}[yshift=-1cm]
  \filldraw[green!80!black] (-3.5, -4.75  ) coordinate (d) circle (2.5pt);
  \filldraw (-2,-6.25) coordinate (da) circle (1.5pt);
  \filldraw (-2,-4.25) coordinate (db) circle (1.5pt);
  \filldraw (0,-5.25) coordinate (dd) circle (1.5pt);
  \filldraw (0,-3.25) coordinate (dc) circle (1.5pt);
  \draw [line width=\linethickness] (d) -- (da);
  \draw [line width=\linethickness] (d) -- (db);
  \draw [line width=\linethickness] (d) -- (dc);
  \draw [line width=\linethickness] (d) -- (dd);
  \draw[->,line width=0.9mm, green!80!black] (d) -- (-3.5,-6.5) node[xshift=-0.5mm, yshift=-3mm] {\huge $\bm{y}$};
  \draw[->,line width=0.9mm, green!80!black] (d) -- (-1,-5) node[xshift=3mm, yshift=-0.8mm] {\huge $\bm{z}$};
  \draw[->,line width=0.9mm, green!80!black] (d) -- (-4.5,-5.5) node[xshift=-1.5mm, yshift=-3mm] {\huge $\bm{x}$};
  \filldraw [line width = \linethicknessframe, fill=green!40, fill opacity=0.4] (da) -- (db) -- (dc) -- (dd) -- (da);
  \end{scope}

  \begin{scope}[xshift=6cm,yshift=1cm]
  \filldraw[red] (-3.5, -4.75  ) coordinate (e) circle (2.5pt);
  \filldraw (-2,-6.25) coordinate (ea) circle (1.5pt);
  \filldraw (-2,-4.25) coordinate (eb) circle (1.5pt);
  \filldraw (0,-5.25) coordinate (ed) circle (1.5pt);
  \filldraw (0,-3.25) coordinate (ec) circle (1.5pt);
  \draw [line width=\linethickness] (e) -- (ea);
  \draw [line width=\linethickness] (e) -- (eb);
  \draw [line width=\linethickness] (e) -- (ec);
  \draw [line width=\linethickness] (e) -- (ed);
  \draw[->,line width=0.9mm, red] (e) -- (-3.5,-6.5) node[xshift=-0.5mm, yshift=-3mm] {\huge $\bm{y}$};
  \draw[->,line width=0.9mm, red] (e) -- (-1,-5) node[xshift=3mm, yshift=-0.8mm] {\huge $\bm{z}$};
  \draw[->,line width=0.9mm, red] (e) -- (-4.5,-5.5) node[xshift=-1.5mm, yshift=-3mm] {\huge $\bm{x}$};
  \filldraw [line width = \linethicknessframe, fill=red!40, fill opacity=0.4] (ea) -- (eb) -- (ec) -- (ed) -- (ea);
  \end{scope}

  \draw[latex-, red!40!black, line width=1.5pt]  (e) to[in=80,out=160] (d) node[midway, above, yshift = -3.35cm,xshift=-4mm] {\LARGE$\mat{t}$};
\end{tikzpicture}
\caption{Utilizing the IMU data it is possible to align the camera views, only leaving an unknown translation~$\mat{t}$.
This assumes that the IMU drift is negligible, which is a realistic assumption if the measurements are not taken too
far apart in time.}
\label{fig:drone}
\end{figure}

There are several benefits of reducing the number of point correspondences used to estimate the motion parameters. In most
cases, this comes at the cost of increased complexity of the system to be solved. Solving systems of polynomial
equations numerically, in a sufficiently fast and robust way, is a challenging task.
One popular method, sometimes referred to as the \emph{action matrix method},
works if there are finitely many solutions~\cite{cox2,moller}. The system of polynomial equations defines
an ideal, for which a Gröbner basis can be computed, leading to an elimination template, where the solutions to the original problem
are obtained by solving an eigenvalue problem~\cite{kukelova2008,byrod-etal-ijcv-2009,bujnak-etal-cvpr-2012}.
This process has been automated by several authors, with the automatic solver by Kukelova~\etal{}~\cite{kukelova2008} as one of the first.
Recent advances use syzygies to make the elimination template smaller~\cite{larsson2016eccv},
as well as discarding spurious solutions by saturating the ideal~\cite{larsson2017ICCV}.
Using Gröbner bases is not the only option; it has for example been shown that other bases can yield better performance~\cite{larsson2018cvpr}, and that overcomplete spanning sets sometimes give better numerical stability~\cite{byrod-etal-ijcv-2009}.
Recently, alternative methods relying on resultants show promising results~\cite{bhayani-etal-cvpr-2020,bhayani-etal-arxiv-2020}.

Apart from adding extra constraints to the motion of the cameras, one may add scene requirements to reduce the number
of necessary point correspondences. A classic example is when the point correspondences lie on a plane, in which
case they are related through an invertible projective transformation known as a homography.
In applications where a planar environment is known to exist, such as indoor environments, the minimal number of
point correspondences for the uncalibrated case is reduced from seven to four, and the corresponding system of equations---known as the
Direct Linear Transform (DLT) equations---is linear in the entries of the homography, and can thus be solved using SVD.

If available, additional input data can be obtained from auxiliary sensors. In this paper we will consider
UAVs equipped with an IMU, from which the gravity direction can be obtained. This,
in turn, is assumed to be aligned with the ground plane normal.

Many commercially available UAVs are equipped with a camera that suffers from radial distortion to some degree.
In order for the pinhole camera model to apply, such distortions must be compensated for, which is usually done
in a pre-calibration process involving a calibration target. In contrast, we investigate a process for onboard
radial distortion auto-calibration, \ie{} a method capable of computing the radial distortion profile (and focal length)
of the optics as well as the motion parameters, without a specific calibration target,
thus eliminating the pre-calibration process. This enables the
user of the UAV to exchange optics, without the need of intermediate calibration procedures, which may not
be feasible without a calibration target. The main contributions of this paper are:
\begin{enumerate}[label={(\roman*)}]
\item a novel polynomial solver for simultaneous estimation of radial distortion profile, focal length and motion parameters, suitable for real-time applications,
\item new insights in how to handle IMU drift, and
\item extensive validation on synthetic and real data on a UAV system demonstrating the applicability of the proposed method.
\end{enumerate}

\section{Related work}
In most Simultaneous Localization and Mapping (SLAM) frameworks
the distortion profile is pre-calibrated using a calibration target. This requires extra off-line
processing, as well as scene requirements. For general scenes, there are
a number of algorithms for simultaneously estimating the distortion profile and the motion
parameters. Some authors propose methods based on large-scale optimization
(bundle adjustment)~\cite{meuel-etal-2016}, while others suggest using polynomial
solvers~\cite{josephson-byrod-cvpr-2009,byrod-etal-bmvc-2009,kukelova-etal-cviu-2010,kukelova-padjla-tpami-2011,bujnak-etal-accv-2011,kukelova-etal-iccv-2013,kuang-etal-cvpr-2014,kukelova2015,larsson-etal-iccv-2017b,pritts2017,pritts2018,larsson-etal-cvpr-2018,valtonenoernhag2020}.
A polynomial solver for the minimal case, \ie{} the smallest number of point correspondences,
is referred to as a minimal solver. There are several
reasons to prefer minimal solvers, as they
accurately encode intrinsic constraints, and transfer such properties to the final solution.
Furthermore, they are suitable for robust estimation frameworks,~\eg{} RANSAC, as the number of necessary
iterations (to obtain an inlier set with a pre-defined probability) is minimized.

There exists a number of different models for estimating the distortion profile. One classic
approach, that is still frequently used in applications is the Brown--Conrady model~\cite{brown1966};
however, although exceptions do exist~\cite{larsson-etal-iccv-2019}, the division model by Fitzgibbon~\cite{fitzgibbon2001} is almost universally used in
the construction of minimal solvers that deal with radial distortion.
One reason for this is that the distortion profile can be accurately
estimated using fewer parameters, which is consistent with the general theory behind minimal solvers.
Other parametric models, recently~\eg{}~\cite{schops-etal-cvpr-2020}, have been proposed, but are not
suitable for minimal solvers for the same reason.

There are several methods that leverage the IMU data---or, simply, rely on the mechanical setup to be accurate enough---to assume a
motion model with a known reference direction~\cite{fraundorfer-etal-eccv-2010,naroditsky-etal-tpami-2011,kalantari-etal-jmiv-2011,saurer-iros-2012,sweeney-etal-3dv-2014,saurer-etal-2017,guan-etal-icra-2018,guan-etal-cviu-2018,ding-etal-2019-iccv,guan-etal-cvpr-2020,valtonenoernhag-etal-arxiv-2020}.
None of the mentioned papers, however, include simultaneous radial distortion correction, while
only a handful consider the case of unknown focal length~\cite{guan-etal-cviu-2018,ding-etal-2019-iccv,guan-etal-cvpr-2020,valtonenoernhag-etal-arxiv-2020}.
To the best of our knowledge, we propose the first ever simultaneous distortion correction, focal length and motion estimation algorithm utilizing IMU data.

\begin{figure}[h!]
\centering
\begin{tikzpicture}
  \filldraw[green!80!black] (-3.5, -4.75  ) coordinate (d) circle (2.5pt);
  \filldraw (-2,-6.25) coordinate (da) circle (1.5pt);
  \filldraw (-2,-4.25) coordinate (db) circle (1.5pt);
  \filldraw (0,-5.25) coordinate (dd) circle (1.5pt);
  \filldraw (0,-3.25) coordinate (dc) circle (1.5pt);
  \draw [line width=\linethickness] (d) -- (da);
  \draw [line width=\linethickness] (d) -- (db);
  \draw [line width=\linethickness] (d) -- (dc);
  \draw [line width=\linethickness] (d) -- (dd);
  \draw[->,line width=0.9mm, green!80!black] (d) -- (-3.5,-6.5) node[xshift=-0.5mm, yshift=-3mm] {\huge $\bm{y}$};
  \draw[->,line width=0.9mm, green!80!black] (d) -- (-1,-5) node[xshift=3mm, yshift=-0.8mm] {\huge $\bm{z}$};
  \draw[->,line width=0.9mm, green!80!black] (d) -- (-4.5,-5.5) node[xshift=-1.5mm, yshift=-3mm] {\huge $\bm{x}$};
  \filldraw [line width = \linethicknessframe, fill=green!40, fill opacity=0.4] (da) -- (db) -- (dc) -- (dd) -- (da);

  \node[] at (1.62,-3.75) {\Large IMU drift};
    \begin{scope}[yshift=-1.5cm,xshift=-2cm,yscale=1,xscale=-1,rotate=-80,yslant=0.5]
        \draw[rotate=15,fill=green!40,fill opacity=0.4,line width=1pt,yshift=-6cm,xshift=2cm] (0:1cm) arc (0:-180:1cm) -- (-180:1.25cm) -- (-.75,.75) -- (-180:.25cm) -- (-180:.5cm) arc (-180:0:.5cm)--cycle;
    \end{scope}

\end{tikzpicture}
\caption{The pitch and roll angles can be accurately estimated using an IMU; however,
the yaw angle (about the $y$-axis, also the gravity direction) often suffers from a drift that accumulates over time.
By fusing angular velocity and accelerometer measurements, the drift is negligible for small time frames.}
\label{fig:drift}
\end{figure}
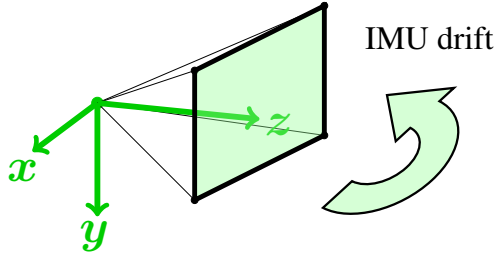

\section{Embracing the IMU drift}\label{sec:embracing}
Prevailing methods have been conceived under the assumption that only two angles can be compensated for using the
IMU data, which is true under general conditions. The drift in the yaw angle, however, is
often very small for consecutive frames. The idea is that we can disregard the error for the yaw
angle initially, and instead correct for it later in the pipeline when enough time has passed
for the drift to make a noticeable impact.
This makes the equations significantly easier to handle, and allows for further intrinsic
calibration, such as radial distortion correction.

Using \cite{5975346}, the orientation can be estimated both robustly and efficiently by fusing angular velocity and accelerometer measurements to form a single estimate of the orientation. The filter operates by integrating gyroscopic data and compensating for bias and integration errors by using the orientation that can be observed from the accelerometer. The rotation about the gravitational direction, however, is not observable and an inevitable drift will accumulate, see~\cref{fig:drift}. The drift is typically very small for short time frames since the sensor noise of the gyroscope is usually very low and the bias changes slowly. 

\subsection{New assumptions on the homography}\label{sec:new assumptions}
Assume the reference direction is known, and aligned with the gravitational
direction, chosen as the~$y$-axis. Then, after a suitable change of coordinates, we may assume that
\begin{equation}\label{eq:Hy}
    \mat{H}_y \sim \mat{I} + \fr{1}{d}\vec{t}\vec{n}^{\T},
\end{equation}
where~$\vec{I}$ is the identity matrix, $\vec{t}$ is the translation vector and~$\mat{n}$
is the unit normal of the plane, see~\cref{fig:drone}. We will assume that the plane normal is aligned with the
gravitational direction, which is a valid assumption when using the ground floor,
thus~$\mat{n}=[0,\,1,\,0]^\T$.
To ease notation, define
\begin{equation}\label{eq:yi}
    \vec{y}_{i}^{(j)} \coloneqq \mat{R}_j^\T\mat{K}^{-1}\vec{x}_i,
\end{equation}
where~$\mat{R}_j$ is the rotation between the two coordinate systems (given by the IMU)
and~$\mat{K} = \diag(f,\,f,\,1)$ is the calibration matrix, where~$f$ is the focal length,
which is assumed to be constant.
Then for two point correspondences~$\vec{x}_1\leftrightarrow\vec{x}_2$ the DLT equations
can be written as
\begin{equation}\label{eq:calib}
    \vec{y}_2^{(2)} \times \mat{H}_y\vec{y}_1^{(1)} = \mat{0}\;.
\end{equation}

The relation between the general (uncalibrated) homography~$\mat{H}$ and $\mat{H}_y$
is thus given by
\begin{equation}\label{eq:HvsHy}
    \mat{H}_y \sim \mat{R}_2^\T\mat{K}^{-1}\mat{H}\mat{K}\mat{R}_1,
\end{equation}
where~$\mat{x}_2\sim\mat{Hx}_1$.
From this, the relative rotation~$\mat{R}_{\te{rel}}$ and the direction of the
relative translation~$\mat{t}_{\te{rel}}$ can be extracted, and are given by
\begin{equation}
    \mat{R}_{\te{rel}} = \mat{R}_2\mat{R}_1^\T
    \quad\te{and}\quad
    \mat{t}_{\te{rel}} \sim \mat{R}_2\mat{t}\;.
\end{equation}
Due to the global scale ambiguity, we may assume $d=1$, and write
\begin{equation}\label{eq:Hyparam}
    \mat{H}_y =
    \begin{bmatrix}
        1 & h_1 & 0 \\
        0 & h_2 & 0 \\
        0 & h_3 & 1
    \end{bmatrix},
\end{equation}
where~$\mat{t}$ can be extracted directly through the entries~$h_i$,
given by
\begin{equation}\label{eq:Ry}
    \mat{t} =
    \begin{bmatrix}
        h_1 \\
        h_2-1 \\
        h_3
    \end{bmatrix}.
\end{equation}

In order to apply the pinhole camera model, radially distorted feature points
must be rectified.
Assuming the distortion can be modeled by the division model~\cite{fitzgibbon2001}, using
only a single distortion parameter~$\lambda$,
the distorted (measured) image point~$\mat{x}_i$ in camera $i$ obeys the relationship
\begin{equation}
\mat{x}_i^u = \phi(\mat{x}_i,\lambda)
= \begin{bmatrix}x_i\\y_i\\1+\lambda(x_i^2+y_i^2)\end{bmatrix},
\end{equation}
where~$\mat{x}_i = [x_i,\, y_i,\, 1]^\T$, and~$\mat{x}_i^u$ are
the undistorted image points compatible with the pinhole camera model.
Here we implicitly assume that the distortion center is at the center of the image.
The modified DLT equations, can therefore be written as
\begin{equation}\label{eq:dlt2}
\phi(\mat{x}_i,\lambda) \times \mat{H} \phi(\mat{x}_j,\lambda) = \mat{0},
\end{equation}
for two point correspondences~$\mat{x}_i\leftrightarrow\mat{x}_j$.

\def\widthfHf{0.3\textwidth}
\def\heightfHf{0.3\textwidth}
\def\widthfrHfr{0.3\textwidth}
\def\heightfrHfr{0.3\textwidth}
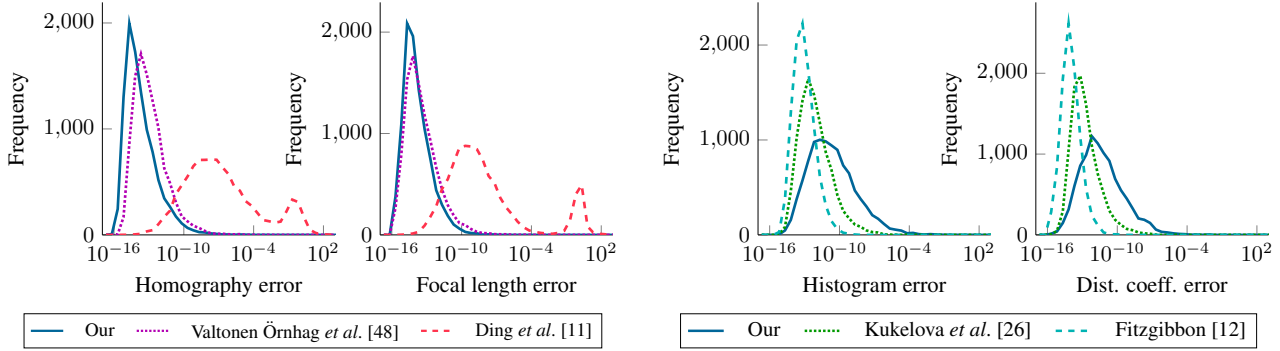
\begin{figure*}[t]
\centering
\begin{tabular}{cc}
\begin{tikzpicture}[>=latex]
    \begin{axis}[
        width=\widthfHf,
        height=\heightfHf,
        axis x line*=bottom,
        axis y line*=left,
        ylabel={Frequency} ,
        ymin = 0,
        xmin = -17,
        xmax = 3,
        ytick = {0, 1000, 2000},
        xtick = {-16, -10, -4, 2},
        xticklabels={$10^{-16}$,$10^{-10}$,$10^{-4}$,$10^{2}$},
        xlabel = {Homography error},
  ]
        \addplot[very thick,color=ourcolor] table {data/H_error_fHf.txt};
        \addplot[very thick,densely dotted,color=valtonenoernhagcolor] table {data/H_error_fHf_icpr_2020.txt};
        \addplot[very thick,dashed,color=dingcolor] table {data/H_error_fHf_ding.txt};
    \end{axis}
    \begin{axis}[
        xshift=.25\textwidth,
        width=\widthfHf,
        height=\heightfHf,
        axis x line*=bottom,
        axis y line*=left,
        ylabel={Frequency} ,
        ymin = 0,
        xmin = -17,
        xmax = 3,
        xtick = {-16, -10, -4, 2},
        xticklabels={$10^{-16}$,$10^{-10}$,$10^{-4}$,$10^{2}$},
        xlabel = {Focal length error},
  ]
        \addplot[very thick,color=ourcolor] table {data/f_error_fHf.txt};
        \addplot[very thick,densely dotted,color=valtonenoernhagcolor] table {data/f_error_fHf_icpr_2020.txt};
        \addplot[very thick,dashed,color=dingcolor] table {data/f_error_fHf_ding.txt};
    \end{axis}
\end{tikzpicture} &%
\begin{tikzpicture}[>=latex]
    \begin{axis}[
        width=\widthfrHfr,
        height=\heightfrHfr,
        axis x line*=bottom,
        axis y line*=left,
        ylabel={Frequency} ,
        ymin = 0,
        xmin = -17,
        xmax = 3,
        xtick = {-16, -10, -4, 2},
        xticklabels={$10^{-16}$,$10^{-10}$,$10^{-4}$,$10^{2}$},
        xlabel={Histogram error}, 
  ]
        \addplot[very thick, color=ourcolor] table {data/H_error_frHfr_elim.txt};
        \addplot[very thick, densely dotted, color=kukelovacolor] table {data/H_error_kukelova.txt};
        \addplot[very thick, dashed, color=fitzgibboncolor] table {data/H_error_fitzgibbon.txt};
    \end{axis}
    \begin{axis}[
        xshift=0.25\textwidth,
        width=\widthfrHfr,
        height=\heightfrHfr,
        axis x line*=bottom,
        axis y line*=left,
        ylabel={Frequency} ,
        ymin = 0,
        xmin = -17,
        xmax = 3,
        xtick = {-16, -10, -4, 2},
        xticklabels={$10^{-16}$,$10^{-10}$,$10^{-4}$,$10^{2}$},
        xlabel={Dist. coeff. error}, 
  ]
        \addplot[very thick, color=ourcolor] table {data/r_error_frHfr_elim.txt};
        \addplot[very thick, densely dotted, color=kukelovacolor] table {data/r_error_kukelova.txt};
        \addplot[very thick, dashed, color=fitzgibboncolor] table {data/r_error_fitzgibbon.txt};
    \end{axis}
\end{tikzpicture} \\%
\includestandalone[height=4.65mm]{./graphs/histogram/histogram_legend_fHf} &%
\includestandalone[height=4.65mm]{./graphs/histogram/histogram_legend_frHfr} %
\end{tabular}
\caption{Error histogram for 10,000 randomly generated problem instances for the proposed solvers: (left) $fHf$ and (right) $frHfr$.
The solver~\cite{kukelova2015} estimates two focal lengths, and we calculate the error for both and
report the geometric mean. Most solvers have an acceptable error distribution, since it in
practice rarely has an impact if the error is of the magnitude $10^{-14}$ or $10^{-10}$. }
\label{fig:synth}
\end{figure*}

\subsection{Benefits of this approach}\label{sec:argument}
The homography described in~\cref{sec:new assumptions} is greatly simplified compared to a general homography and has fewer parameters that need to be determined.
In the case of unknown radial distortion profile, the competing
methods~\cite{kukelova2015,fitzgibbon2001} return a general homography, \ie{} with eight degrees
of freedom. Unless one makes assumptions about the motion of the cameras---for example that it consists only of pure rotations---it is not possible to extract the motion parameters, even in the partially calibrated case.
To see this, note that a Euclidean homography
\begin{equation}
    \mat{H}_{\te{euc}} \sim \mat{R} + \mat{tn}^\T,
\end{equation}
has eight degrees of freedom---three in~$\mat{R}$, three in~$\mat{t}$ and two in~$\mat{n}$ (since the
length of~$\mat{n}$ is arbitrary). This is to be compared to a general homography that also has
eight degrees of freedom. We conclude that a partially calibrated homography on the form~$\mat{K}\mat{H}_{\te{euc}}\mat{K}^{-1}$ must have nine degrees of freedom (the focal length~$f$ in~$\mat{K}$ and the eight from $\mat{H}_{\te{euc}}$), hence is over-parametrized,~\ie{}
there exists a one-dimensional family of possible decompositions.
For this reason, we cannot extract the pose of the methods~\cite{kukelova2015,fitzgibbon2001}, unless
we assume that we know the focal length~\emph{a~priori}, or constrain the motion.
This, in itself, makes the methods infeasible to include
in a SLAM framework, where we want to estimate the camera positions.

\section{Polynomial solvers}
In this section we present two-sided solvers, \ie{} when the
same intrinsic parameters (focal length and/or radial distortion parameter)
are assumed for both cameras.

\subsection{Calibrated case (1.5 point)}\label{sec:calibrated}
This case does not model an unknown focal length or distortion parameter,
and is essentially the same approach as in~\cite{guan-etal-icra-2018}, but is given here for completeness.
Given 1.5 point correspondences it is possible to form the
linear system~$\mat{A}\vec{h} = \vec{b}$,
where $\mat{A}$ is a $3\times 3$ matrix and~$\vec{h}$ contains the~$h_i$ from~\eqref{eq:Hyparam}.
For non-degenerate configurations, the matrix~$\mat{A}$ has full rank, and the solution can
be obtained immediately as $\vec{h}=\mat{A}^{-1}\vec{b}$.
This is a very fast solver, since it is linear and can be solved without SVD.

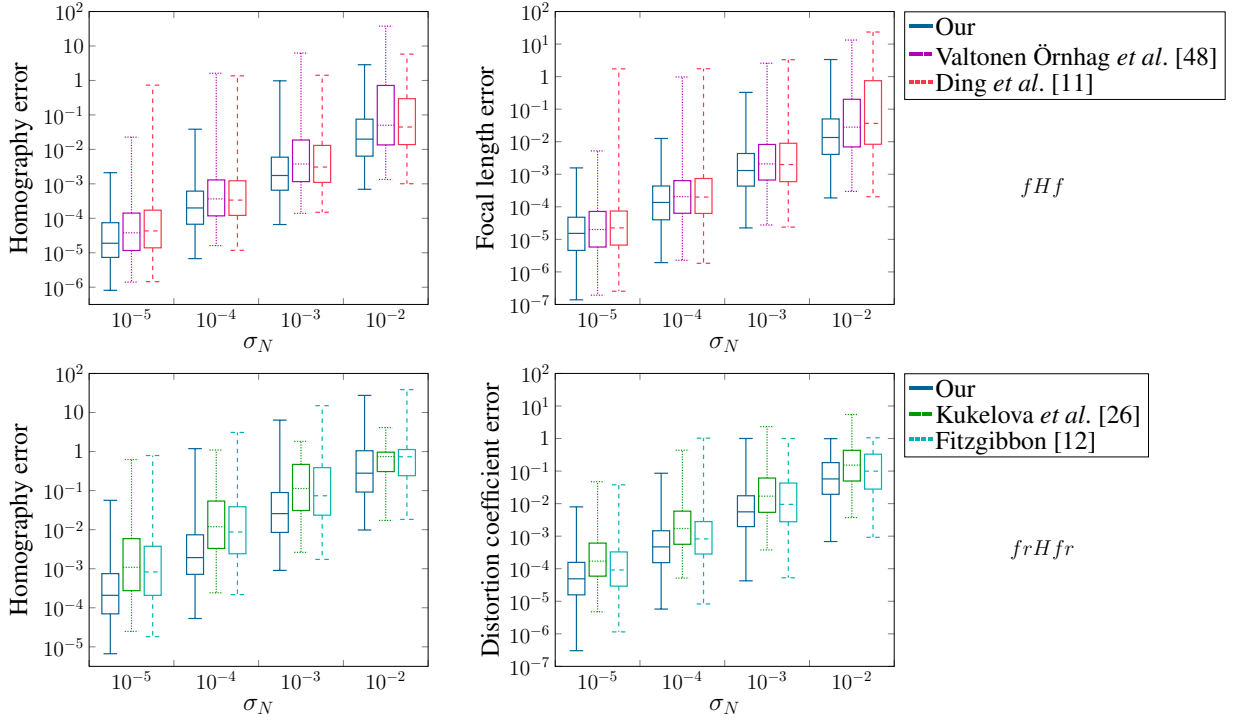
\begin{figure*}[t!]
\centering
\begin{minipage}{0.88\linewidth}
\begin{tikzpicture}
\begin{axis}[
boxplot/draw direction=y,
ylabel={Homography error},
xlabel={$\sigma_N$},
label style={font=\LARGE},
height=\boxplotheight,
width=\boxplotwidth,
xmin=0,xmax=4,
ymin=-6.5,ymax=2,
cycle list={{ourcolor, \thicknesslineplot},{valtonenoernhagcolor, densely dotted, \thicknesslineplot}, {dingcolor, dashed, \thicknesslineplot}},
boxplot/every box/.style={solid},
boxplot prepared={
        %
        draw position={1/4 + floor(\plotnumofactualtype/3) + 1/4*fpumod(\plotnumofactualtype,3)},
        %
        box extend=0.2,
},
x=2cm,
xtick={0,1,2,...,4},
x tick label as interval,
xticklabels={%
        {$10^{-5}$},%
        {$10^{-4}$},%
        {$10^{-3}$},%
        {$10^{-2}$},%
},
x tick label style={
        text width=2.5cm,
        align=center,
        font=\Large
},
ytick={-6,-5,...,2},
yticklabels={%
        {$10^{-6}$},%
        {$10^{-5}$},%
        {$10^{-4}$},%
        {$10^{-3}$},%
        {$10^{-2}$},%
        {$10^{-1}$},%
        {$1$},%
        {$10$},%
        {$10^{2}$},%
},
y tick label style={
        font=\Large,
},
]
HOMOGRAPHY ERROR
\addplot+[boxplot prepared={
	 upper whisker=-2.6763,
	 upper quartile=-4.1277,
	 median=-4.7243,
	 lower quartile=-5.1354,
	 lower whisker=-6.0903,
}] coordinates {};

\addplot+[boxplot prepared={
	 upper whisker=-1.6450,
	 upper quartile=-3.8493,
	 median=-4.4215,
	 lower quartile=-4.9340,
	 lower whisker=-5.8528,
}] coordinates {};

\addplot+[boxplot prepared={
	 upper whisker=-0.1365,
	 upper quartile=-3.7642,
	 median=-4.3664,
	 lower quartile=-4.8572,
	 lower whisker=-5.8391,
}] coordinates {};

\addplot+[boxplot prepared={
	 upper whisker=-1.4147,
	 upper quartile=-3.2119,
	 median=-3.6999,
	 lower quartile=-4.1726,
	 lower whisker=-5.1703,
}] coordinates {};

\addplot+[boxplot prepared={
	 upper whisker=0.2096,
	 upper quartile=-2.8860,
	 median=-3.4366,
	 lower quartile=-3.9293,
	 lower whisker=-4.7926,
}] coordinates {};

\addplot+[boxplot prepared={
	 upper whisker=0.1344,
	 upper quartile=-2.9112,
	 median=-3.4720,
	 lower quartile=-3.9162,
	 lower whisker=-4.9297,
}] coordinates {};

\addplot+[boxplot prepared={
	 upper whisker=-0.0102,
	 upper quartile=-2.2263,
	 median=-2.7582,
	 lower quartile=-3.1850,
	 lower whisker=-4.1811,
}] coordinates {};

\addplot+[boxplot prepared={
	 upper whisker=0.7960,
	 upper quartile=-1.7283,
	 median=-2.4249,
	 lower quartile=-2.9358,
	 lower whisker=-3.8573,
}] coordinates {};

\addplot+[boxplot prepared={
	 upper whisker=0.1502,
	 upper quartile=-1.8846,
	 median=-2.5133,
	 lower quartile=-2.9578,
	 lower whisker=-3.8251,
}] coordinates {};

\addplot+[boxplot prepared={
	 upper whisker=0.4567,
	 upper quartile=-1.1233,
	 median=-1.7031,
	 lower quartile=-2.1968,
	 lower whisker=-3.1585,
}] coordinates {};

\addplot+[boxplot prepared={
	 upper whisker=1.5756,
	 upper quartile=-0.1450,
	 median=-1.2997,
	 lower quartile=-1.8703,
	 lower whisker=-2.8744,
}] coordinates {};

\addplot+[boxplot prepared={
	 upper whisker=0.7637,
	 upper quartile=-0.5294,
	 median=-1.3516,
	 lower quartile=-1.8615,
	 lower whisker=-2.9964,
}] coordinates {};
\end{axis}
\begin{axis}[
xshift=11cm,
legend cell align=left,
legend pos=outer north east,
legend entries = {Our, Valtonen Örnhag~\etal{}~\cite{valtonenoernhag-etal-arxiv-2020}, Ding~\etal{}~\cite{ding-etal-2019-iccv}},
legend style={font=\LARGE},
boxplot/draw direction=y,
ylabel={Focal length error},
xlabel={$\sigma_N$},
label style={font=\LARGE},
height=\boxplotheight,
width=\boxplotwidth,
xmin=0,xmax=4,
ymin=-7,ymax=2,
cycle list={{ourcolor, \thicknesslineplot},{valtonenoernhagcolor, densely dotted, \thicknesslineplot}, {dingcolor, dashed, \thicknesslineplot}},
boxplot/every box/.style={solid},
boxplot prepared={
        %
        draw position={1/4 + floor(\plotnumofactualtype/3) + 1/4*fpumod(\plotnumofactualtype,3)},
        %
        box extend=0.2,
},
x=2cm,
xtick={0,1,2,...,4},
x tick label as interval,
xticklabels={%
        {$10^{-5}$},%
        {$10^{-4}$},%
        {$10^{-3}$},%
        {$10^{-2}$},%
},
x tick label style={
        text width=2.5cm,
        align=center,
        font=\Large
},
ytick={-7,-6,...,2},
yticklabels={%
        {$10^{-7}$},%
        {$10^{-6}$},%
        {$10^{-5}$},%
        {$10^{-4}$},%
        {$10^{-3}$},%
        {$10^{-2}$},%
        {$10^{-1}$},%
        {$1$},%
        {$10^{1}$},%
        {$10^{2}$},%
        {$10^{3}$},%
},
y tick label style={
        font=\Large,
},
]
\addlegendimage{legend image with solid};
\addlegendimage{legend image with dots2};
\addlegendimage{legend image with stripes2};
\addplot+[boxplot prepared={
	 upper whisker=-2.8043,
	 upper quartile=-4.3205,
	 median=-4.8167,
	 lower quartile=-5.3402,
	 lower whisker=-6.8577,
}] coordinates {};

\addplot+[boxplot prepared={
	 upper whisker=-2.2826,
	 upper quartile=-4.1424,
	 median=-4.6978,
	 lower quartile=-5.2356,
	 lower whisker=-6.7154,
}] coordinates {};

\addplot+[boxplot prepared={
	 upper whisker=0.2384,
	 upper quartile=-4.1304,
	 median=-4.6468,
	 lower quartile=-5.1765,
	 lower whisker=-6.5940,
}] coordinates {};

\addplot+[boxplot prepared={
	 upper whisker=-1.9004,
	 upper quartile=-3.3608,
	 median=-3.8661,
	 lower quartile=-4.3990,
	 lower whisker=-5.7143,
}] coordinates {};

\addplot+[boxplot prepared={
	 upper whisker=-0.0099,
	 upper quartile=-3.1963,
	 median=-3.6876,
	 lower quartile=-4.1977,
	 lower whisker=-5.6433,
}] coordinates {};

\addplot+[boxplot prepared={
	 upper whisker=0.2425,
	 upper quartile=-3.1305,
	 median=-3.7000,
	 lower quartile=-4.2041,
	 lower whisker=-5.7358,
}] coordinates {};

\addplot+[boxplot prepared={
	 upper whisker=-0.4848,
	 upper quartile=-2.3622,
	 median=-2.8852,
	 lower quartile=-3.3641,
	 lower whisker=-4.6483,
}] coordinates {};

\addplot+[boxplot prepared={
	 upper whisker=0.4121,
	 upper quartile=-2.0838,
	 median=-2.6802,
	 lower quartile=-3.1770,
	 lower whisker=-4.5588,
}] coordinates {};

\addplot+[boxplot prepared={
	 upper whisker=0.5196,
	 upper quartile=-2.0482,
	 median=-2.7016,
	 lower quartile=-3.2260,
	 lower whisker=-4.6243,
}] coordinates {};

\addplot+[boxplot prepared={
	 upper whisker=0.5251,
	 upper quartile=-1.3003,
	 median=-1.8706,
	 lower quartile=-2.3898,
	 lower whisker=-3.7271,
}] coordinates {};

\addplot+[boxplot prepared={
	 upper whisker=1.1245,
	 upper quartile=-0.6964,
	 median=-1.5521,
	 lower quartile=-2.1602,
	 lower whisker=-3.5273,
}] coordinates {};

\addplot+[boxplot prepared={
	 upper whisker=1.3702,
	 upper quartile=-0.1249,
	 median=-1.4361,
	 lower quartile=-2.0786,
	 lower whisker=-3.6915,
}] coordinates {};
\end{axis}

\node[] at (22.5,2.75) {\Large $fHf$};

\end{tikzpicture}\\%
\begin{tikzpicture}
\begin{axis}[
boxplot/draw direction=y,
ylabel={Homography error},
xlabel={$\sigma_N$},
label style={font=\LARGE},
height=\boxplotheight,
width=\boxplotwidth,
xmin=0,xmax=4,
ymin=-5.5,ymax=2,
cycle list={{ourcolor, \thicknesslineplot},{kukelovacolor, densely dotted, \thicknesslineplot},{fitzgibboncolor, dashed, \thicknesslineplot}},
boxplot/every box/.style={solid},
boxplot prepared={
        %
        draw position={1/4 + floor(\plotnumofactualtype/3) + 1/4*fpumod(\plotnumofactualtype,3)},
        %
        box extend=0.2,
},
x=2cm,
xtick={0,1,2,...,4},
x tick label as interval,
xticklabels={%
        {$10^{-5}$},%
        {$10^{-4}$},%
        {$10^{-3}$},%
        {$10^{-2}$},%
},
x tick label style={
        text width=2.5cm,
        align=center,
        font=\Large
},
ytick={-7,-6,...,2},
yticklabels={%
        {$10^{-7}$},%
        {$10^{-6}$},%
        {$10^{-5}$},%
        {$10^{-4}$},%
        {$10^{-3}$},%
        {$10^{-2}$},%
        {$10^{-1}$},%
        {$1$},%
        {$10$},%
        {$10^{2}$},%
},
y tick label style={
        font=\Large,
},
]
\addplot+[boxplot prepared={
	 upper whisker=-1.2486,
	 upper quartile=-3.1252,
	 median=-3.6809,
	 lower quartile=-4.1557,
	 lower whisker=-5.1768,
}] coordinates {};

\addplot+[boxplot prepared={
	 upper whisker=-0.2040,
	 upper quartile=-2.2297,
	 median=-2.9638,
	 lower quartile=-3.5615,
	 lower whisker=-4.6062,
}] coordinates {};

\addplot+[boxplot prepared={
	 upper whisker=-0.1020,
	 upper quartile=-2.4249,
	 median=-3.0816,
	 lower quartile=-3.6813,
	 lower whisker=-4.7399,
}] coordinates {};

\addplot+[boxplot prepared={
	 upper whisker=0.0749,
	 upper quartile=-2.1313,
	 median=-2.7152,
	 lower quartile=-3.1437,
	 lower whisker=-4.2747,
}] coordinates {};

\addplot+[boxplot prepared={
	 upper whisker=0.0389,
	 upper quartile=-1.2678,
	 median=-1.9245,
	 lower quartile=-2.4805,
	 lower whisker=-3.6167,
}] coordinates {};

\addplot+[boxplot prepared={
	 upper whisker=0.4896,
	 upper quartile=-1.4135,
	 median=-2.0603,
	 lower quartile=-2.6163,
	 lower whisker=-3.6629,
}] coordinates {};

\addplot+[boxplot prepared={
	 upper whisker=0.8052,
	 upper quartile=-1.0490,
	 median=-1.5885,
	 lower quartile=-2.0710,
	 lower whisker=-3.0436,
}] coordinates {};

\addplot+[boxplot prepared={
	 upper whisker=0.2614,
	 upper quartile=-0.3289,
	 median=-0.9459,
	 lower quartile=-1.5091,
	 lower whisker=-2.5809,
}] coordinates {};

\addplot+[boxplot prepared={
	 upper whisker=1.1746,
	 upper quartile=-0.4119,
	 median=-1.1316,
	 lower quartile=-1.6290,
	 lower whisker=-2.7617,
}] coordinates {};

\addplot+[boxplot prepared={
	 upper whisker=1.4387,
	 upper quartile=0.0213,
	 median=-0.5536,
	 lower quartile=-1.0370,
	 lower whisker=-2.0082,
}] coordinates {};

\addplot+[boxplot prepared={
	 upper whisker=0.6129,
	 upper quartile=-0.0158,
	 median=-0.1280,
	 lower quartile=-0.5119,
	 lower whisker=-1.7637,
}] coordinates {};

\addplot+[boxplot prepared={
	 upper whisker=1.5858,
	 upper quartile=0.0524,
	 median=-0.1335,
	 lower quartile=-0.6179,
	 lower whisker=-1.7378,
}] coordinates {};
\end{axis}
\begin{axis}[
xshift=11cm,
legend cell align=left,
legend pos=outer north east,
legend entries = {Our, Kukelova~\etal{}~\cite{kukelova2015}, Fitzgibbon~\cite{fitzgibbon2001}},
legend style={font=\LARGE},
boxplot/draw direction=y,
ylabel={Distortion coefficient error},
xlabel={$\sigma_N$},
label style={font=\LARGE},
height=\boxplotheight,
width=\boxplotwidth,
xmin=0,xmax=4,
ymin=-7,ymax=2,
cycle list={{ourcolor, \thicknesslineplot},{kukelovacolor, densely dotted, \thicknesslineplot},{fitzgibboncolor, dashed, \thicknesslineplot}},
boxplot/every box/.style={solid},
boxplot prepared={
        %
        draw position={1/4 + floor(\plotnumofactualtype/3) + 1/4*fpumod(\plotnumofactualtype,3)},
        %
        box extend=0.2,
},
x=2cm,
xtick={0,1,2,...,4},
x tick label as interval,
xticklabels={%
        {$10^{-5}$},%
        {$10^{-4}$},%
        {$10^{-3}$},%
        {$10^{-2}$},%
},
x tick label style={
        text width=2.5cm,
        align=center,
        font=\Large
},
ytick={-7,-6,...,2},
yticklabels={%
        {$10^{-7}$},%
        {$10^{-6}$},%
        {$10^{-5}$},%
        {$10^{-4}$},%
        {$10^{-3}$},%
        {$10^{-2}$},%
        {$10^{-1}$},%
        {$1$},%
        {$10^{1}$},%
        {$10^{2}$},%
},
y tick label style={
        font=\Large,
},
]
\addlegendimage{legend image with solid};
\addlegendimage{legend image with dots};
\addlegendimage{legend image with stripes};
\addplot+[boxplot prepared={
	 upper whisker=-6.5196,
	 upper quartile=-4.8027,
	 median=-4.3102,
	 lower quartile=-3.8053,
	 lower whisker=-2.0988,
}] coordinates {};

\addplot+[boxplot prepared={
	 upper whisker=-1.3262,
	 upper quartile=-3.2131,
	 median=-3.7671,
	 lower quartile=-4.2283,
	 lower whisker=-5.3237,
}] coordinates {};

\addplot+[boxplot prepared={
	 upper whisker=-5.9406,
	 upper quartile=-4.5348,
	 median=-4.0383,
	 lower quartile=-3.4835,
	 lower whisker=-1.4221,
}] coordinates {};

\addplot+[boxplot prepared={
	 upper whisker=-5.2401,
	 upper quartile=-3.8133,
	 median=-3.3294,
	 lower quartile=-2.8334,
	 lower whisker=-1.0667,
}] coordinates {};

\addplot+[boxplot prepared={
	 upper whisker=-0.3595,
	 upper quartile=-2.2314,
	 median=-2.7638,
	 lower quartile=-3.2491,
	 lower whisker=-4.2885,
}] coordinates {};

\addplot+[boxplot prepared={
	 upper whisker=-5.0815,
	 upper quartile=-3.5478,
	 median=-3.0832,
	 lower quartile=-2.5517,
	 lower whisker=0.0123,
}] coordinates {};

\addplot+[boxplot prepared={
	 upper whisker=-4.3723,
	 upper quartile=-2.7056,
	 median=-2.2525,
	 lower quartile=-1.7582,
	 lower whisker=0.0016,
}] coordinates {};

\addplot+[boxplot prepared={
	 upper whisker=0.3653,
	 upper quartile=-1.2139,
	 median=-1.7706,
	 lower quartile=-2.2673,
	 lower whisker=-3.4228,
}] coordinates {};

\addplot+[boxplot prepared={
	 upper whisker=-4.2781,
	 upper quartile=-2.5568,
	 median=-2.0228,
	 lower quartile=-1.3682,
	 lower whisker=0.0012,
}] coordinates {};

\addplot+[boxplot prepared={
	 upper whisker=-3.1647,
	 upper quartile=-1.7136,
	 median=-1.2409,
	 lower quartile=-0.7396,
	 lower whisker=-0.0039,
}] coordinates {};

\addplot+[boxplot prepared={
	 upper whisker=0.7385,
	 upper quartile=-0.3620,
	 median=-0.8160,
	 lower quartile=-1.3047,
	 lower whisker=-2.4280,
}] coordinates {};

\addplot+[boxplot prepared={
	 upper whisker=-3.0355,
	 upper quartile=-1.5536,
	 median=-1.0019,
	 lower quartile=-0.4808,
	 lower whisker=0.0229,
}] coordinates {};
\end{axis}

\node[] at (22.5,2.75) {\Large $frHfr$};

\end{tikzpicture}
\end{minipage}
\caption{Noise sensitivity comparison for Gaussian noise with standard deviation~$\sigma_N$.
For each noise level 1,000 random problem instances were generated.
The geometric mean error is shown for~\cite{kukelova2015}.}
\label{fig:noise}
\end{figure*}

\subsection{Equal and unknown focal length (\texorpdfstring{$fHf$}{fHf}, 2 point)}\label{sec:fHf}
Parameterize the inverse of the unknown calibration matrix
as~$\mat{K}^{-1} = \diag(1,\,1,\,w)$, and consider the
rectified points~\eqref{eq:yi}, which now depend linearly on the unknown parameter~$w$.
Parameterizing~$\Hy$ as in~\eqref{eq:Hyparam}, it is clear that
the equations obtained from~\eqref{eq:calib} are linear in $h_1$, $h_2$ and $h_3$ and quadratic
in~$w$. This system of equations has infinitely many solutions, if we allow~\mbox{$w=0$}.
Such solutions, however, do not yield geometrically meaningful reconstructions, and should
therefore be excluded. This can be achieved using saturation, through the method suggested
in~\cite{larsson2017ICCV}.

We exploit the linear relation of~$h_1$, $h_2$, $h_3$, making it
possible to write the equations as
\begin{equation}
    \mat{M}\begin{bmatrix}
    \mat{h} \\ 1
    \end{bmatrix}
    = \mat{0},
\end{equation}
where~$\mat{M}$ is a $4\times 4$ matrix depending on~$w$, and~$\mat{h}$ is the vector
containing the elements~$h_i$. Thus, we may consider finding
a non-trivial nullspace of~$\mat{M}$, which exists if and only if~$\det{\mat{M}}=0$.
This equation reduces to a sextic polynomial in the unknown~$w$, thus has six solutions, which
can be found using a simple root finding algorithm (action matrix method is not necessary).
Since we know from before that the original problem has four solutions, we conclude that
two spurious solutions have been added; in fact, these can easily be disregarded as a pre-processing
step, as they correspond to nullspace basis vectors with last element equal to zero.
Numerical tests confirm that this is the case.

When the (up to) four possible real solutions of~$w$ have been obtained, the unknowns
$h_1$, $h_2$ and $h_3$ can be obtained using SVD.

\subsection{Equal and unknown focal length and radial distortion coefficient(\texorpdfstring{$frHfr$}{frHfr}, 2.5-point)}\label{sec:frHfr}
Let us now consider the case with equal and unknown focal length
and radial distortion coefficient. We use the division model introduced in~\cite{fitzgibbon2001},
with a single distortion parameter~$\lambda$.

Given two point correspondences~$\mat{x}_1\leftrightarrow\mat{x}_2$, the modified DLT
equations~\eqref{eq:dlt2} hold true.
Building an elimination template from these equations yields a large and numerically unstable
solver, and therefore, we reparameterize the problem.
Applying~$\mat{H}_y=\mat{I}+\mat{tn}^\T$ to~\eqref{eq:HvsHy}, we get
\begin{equation}
    \mat{K}^{-1}\mat{HK} \sim \mat{R}_2\mat{H}_y\mat{R}_1^\T = \mat{R}_2\mat{R}_1^\T + \mat{R}_2\mat{t}\mat{n}^T\mat{R}_1^\T\;.
\end{equation}
Introducing $\hat{\mat{R}} = \mat{R}_2\mat{R}_1^\T$, $\hat{\mat{t}} = \mat{R}_2\mat{t}$ and
$\hat{\mat{n}} = \mat{R}_1\mat{n} = [r_{12},\,r_{22},\,r_{32}]^\T$, the general homography
can now be written as
\begin{equation}\label{eq:rewrite}
    \mat{H} \sim\mat{K}\left( \hat{\mat{R}}+\hat{\mat{t}}\hat{\mat{n}}^T\right)\mat{K}^{-1}\;.
\end{equation}
This accomplishes two things: (1) we have replaced several multiplications, (2) we have reduced
the number of input data necessary. Analyzing the quotient ring of the corresponding ideal,
we conclude that there are three possible solutions.

\def\stitchingimagewidth{0.318\linewidth}
\begin{figure*}[t]
    \centering
    \begin{subfigure}[b]{\stitchingimagewidth}
        \centering
        \includegraphics[width=\textwidth]{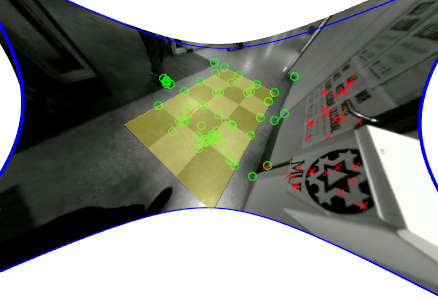}
        \caption{Our, 70 \% inliers.}
    \end{subfigure}
    \begin{subfigure}[b]{\stitchingimagewidth}
        \centering
        \includegraphics[width=\textwidth]{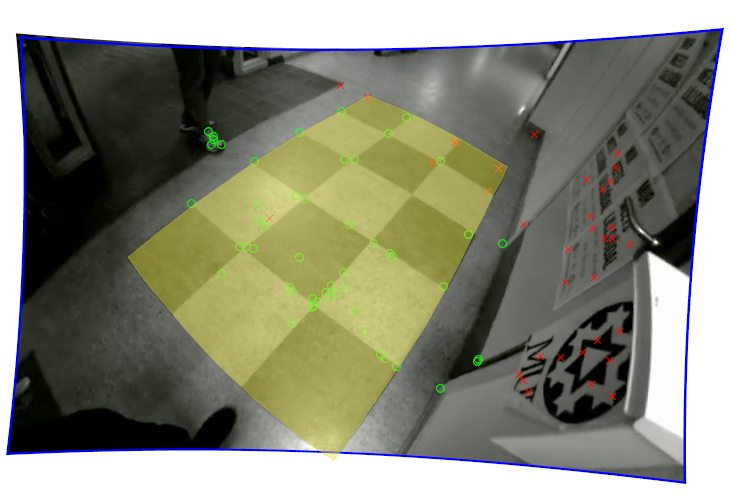}
        \caption{Kukelova~\etal{}~\cite{kukelova2015}, 62 \% inliers.}
    \end{subfigure}
    \begin{subfigure}[b]{\stitchingimagewidth}
        \centering
        \includegraphics[width=\textwidth]{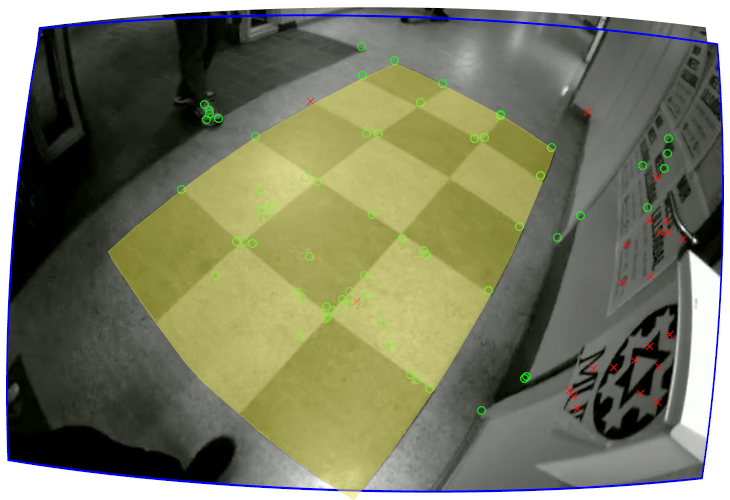}
        \caption{Fitzgibbon~\cite{fitzgibbon2001}, 67 \% inliers.}
    \end{subfigure}
    \vspace{-2mm}
    \caption{Selection of panoramas created with the competing methods. The blue
             frame is added for visualization, as well as inliers (green circles) and outliers
             (red crosses). Note that none of the methods require a checkerboard to be visible
             in the scene, but is simply chosen to ease the ocular inspection of the stitching.
             A correct rectification will map physically straight lines to straight lines,~\ie{}
             the yellow area should be a quadrilateral. Only our method is capable of producing
             this result.
             }
    \label{fig:ransac_visual}
\end{figure*}

We parameterize the calibration matrix as~$\mat{K}=\diag(f,\,f,\,1)$
and its inverse~$\mat{K}^{-1}=\diag(f^{-1},\,f^{-1},\,1)$, respectively.
From here on it would be possible to construct an elimination template; however,
we may eliminate one variable in order to get a reduced system.
Using only the third row of~\eqref{eq:dlt2}, from three point correspondences,
one obtains a system on the form \mbox{$\mat{Mv}=\mat{0}$},
where $\mat{M}$ is a $3\times 9$ coefficient matrix, and~$\mat{v}$
is the vector of monomials, more precisely\footnote{It turns out that the third row does not
contain any reciprocal~$f$.}
\begin{equation}
    \vec{v} = \begin{bmatrix}
        \hat{t}_1f\lambda &
        \hat{t}_1f &
        \hat{t}_1 &
        \hat{t}_2f\lambda &
        \hat{t}_2f &
        \hat{t}_2 &
        f\lambda &
        f &
        1
    \end{bmatrix}^\T\;.
\end{equation}
Since~$\hat{t}_1$ and $\hat{t}_2$ are present in only three monomials, either of the
two can be eliminated; we will proceed by eliminating the latter, as it yields a smaller elimination
template.
After Gauss--Jordan elimination, the coefficient matrix is given by
\setcounter{MaxMatrixCols}{20}
\begin{equation}
\raisebox{-6pt}{ $\hat{\mat{M}} =$ }
\raisebox{-6pt}{\HUGE[}
\begingroup 
\setlength\arraycolsep{2pt}
\begin{matrix}
        \hat{t}_2f\lambda &
        \hat{t}_2f &
        \hat{t}_2 &
        \hat{t}_1f\lambda &
        \hat{t}_1f &
        \hat{t}_1 &
        f\lambda &
        f &
        1\\
    1 & & & \bullet & \bullet & \bullet   & \bullet & \bullet & \bullet \\
    & 1 & & \bullet & \bullet & \bullet   & \bullet & \bullet & \bullet \\
    & & 1 & \bullet & \bullet & \bullet   & \bullet & \bullet & \bullet \\
\end{matrix}
\endgroup
\raisebox{-6pt}{\HUGE],}
\end{equation}
from which we establish the following relations
\begin{equation}
\begin{aligned}
    \hat{t}_2f\lambda + g_1(\hat{t}_1,f,\lambda) &= 0, \\
    \hat{t}_2f + g_2(\hat{t}_1,f,\lambda) &= 0, \\
    \hat{t}_2 + g_3(\hat{t}_1,f,\lambda) &= 0,
\end{aligned}
\end{equation}
where $g_i$ are polynomials of three variables. Furthermore,
the following constraints must be fulfilled
\begin{equation}\label{eq:new_eqs}
\begin{aligned}
    g_1(\hat{t}_2,f,\lambda) - \lambda g_2(\hat{t}_2,f,\lambda) &= 0, \\
    g_2(\hat{t}_2,f,\lambda) - f g_3(\hat{t}_2,f,\lambda) &= 0, \\
    g_1(\hat{t}_2,f,\lambda) - \lambda f g_3(\hat{t}_2,f,\lambda) &= 0\;.
\end{aligned}
\end{equation}
We can now use the first row of~\eqref{eq:dlt2}, from which we get two equations
(which must be multiplied by~$f$ to make it polynomial).
Together with~\eqref{eq:new_eqs} we have five equations in four unknowns.

To build a solver we saturate~$f$, to remove spurious solutions corresponding to zero
focal length. Analyzing the quotient ring we have again three solutions
and the basis heuristic~\cite{larsson2018cvpr} yields a template size of~$26\times 29$.
Using the hidden variable trick, as in~\cref{sec:fHf}, we were able to construct a solver
with a template size of $17\times 20$; however, this solver was not as numerically
stable as the one proposed, nor faster.

\section{Experiments}
\subsection{Numerical stability and noise sensitivity}
We compare the proposed methods with other state-of-the-art methods on synthetic
data, to evaluate the numerical stability.
For the case of unknown focal length we compare to the
2.5~point method~\cite{valtonenoernhag-etal-arxiv-2020} and the 3.5~pt
method~\cite{ding-etal-2019-iccv}, and in the case of unknown radial distortion we
compare to the 5 point methods~\cite{kukelova2015,fitzgibbon2001}.
We generate noise free problem instances, by generating homographies and rotation matrices,
and project a random set of points to establish point correspondences. In the case of radial
distortion, these points are distorted using the division model.
The error histograms are shown in see~\cref{fig:synth}. In the case of unknown focal length
our method is superior to the others; however, with unknown radial distortion, we are not
as stable as others. The accuracy, however, is in the order of $10^{-10}$. This is sufficient
for most applications.
We will see in future experiments, that this does not cause any practical issues.
The homography error is measured as the difference between the estimated homography and the ground truth in the Frobenius norm, normalized with the Frobenius norm
of the ground truth homography, where the homographies are chosen such that~$h_{33}=1$.
The errors of the focal length and radial distortion coefficient are measured as
the absolute difference divided by the ground truth value.

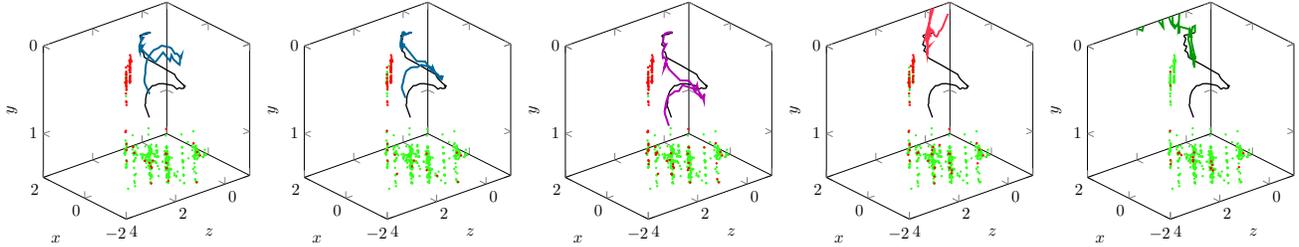
\begin{figure*}[t]
\centering
\pgfplotsset{
    xlabel=$x$,
    ylabel=$z$,
    zlabel=$y$,
    z dir=reverse,
    unit vector ratio*=1 1 3,
    xmin = -2,
    xmax = 2,
    ymin = -1,
    ymax = 4,
    zmin = 0,
    zmax = 1.5,
    view={-130}{25}
}

\newcommand{\trajplot}[2]{
    \begin{tikzpicture}
    \begin{axis}
    \addplot3[inliercolor, mark=*, only marks, mark size=0.4pt] table {../graphs/real_data/inliers_map_2020_09_21_20_52_11_#1.txt};
    \addplot3[outliercolor, mark=*, only marks, mark size=0.4pt] table {../graphs/real_data/outliers_map_2020_09_21_20_52_11_#1.txt};
    \addplot3[gtcolor, thick, mark = none] table {../graphs/real_data/traj_map_2020_09_21_20_52_11_gt.txt};
    \addplot3[#2, mark = none, \thicknesslineplotThree] table {../graphs/real_data/traj_map_2020_09_21_20_52_11_#1.txt};
    \end{axis}
    \end{tikzpicture}
}

\trajplot{fHf}{ourcolor}
\trajplot{frHfr}{ourcolor}
\trajplot{icpr}{valtonenoernhagcolor}
\trajplot{ding}{dingcolor}
\trajplot{kukelova}{kukelovacolor}

\pgfplotsset{}
\caption{Estimated trajectories from the \emph{Indoor} dataset. From left to right:
Our $fHf$, our $frHfr$, Valtonen Örnhag~\etal{}~\cite{valtonenoernhag-etal-arxiv-2020},
Ding~\etal{}~\cite{ding-etal-2019-iccv} and Kukelova~\etal{}~\cite{kukelova-etal-2017-cvpr}.
The green dots indicate points that have been selected as inliers at least once, and the red
points those which have been consistently rejected. Rectified images were used as input to all
solvers except for~$frHfr$, which received the raw input images.}
\label{fig:real_traj}
\end{figure*}

Lastly,
Gaussian noise is added to the image correspondences, in order to compare the noise sensitivity
of the methods. The standard deviation~$\sigma_N$ is varied for a number of different noise
levels. For all noise levels, our solvers perform superior to the other methods,
for both the case with and without radial distortion, see~\cref{fig:noise}.

\subsection{Speed evaluation}\label{sec:timings}
Next we compare the execution time for the considered methods.
We compare the mean execution time given a minimal set of point correspondences until the
putative homographies, and other parameters are obtained,~\ie{} including all pre-processing and
post-processing steps. Furthermore, for the 2.5 and 3.5 point methods, we discard false
solutions using the previously unused DLT equation.

As we are interested in performing the computations onboard the UAV, we evaluate the
performance on a Raspberry Pi~4, and the mean execution times are listed in~\cref{tab:exectime}.
All solvers are implemented in C++ using Eigen~\cite{eigen} and compiled in \texttt{g++}
with the~\texttt{-O2} optimization flag.
Lastly, we list the maximal number of iterations possible on a 30 fps system, which
we will use in~\cref{sec:realdata} to compare real-time performance.

\begin{table}[htb]
\centering
\caption{Mean execution time on a Raspberry Pi~4 for 100,000 randomly generated problem instances in C++.
The last column is the maximal number of iterations possible when running 30 fps.}
\begin{tabular}{lll}
\hline
Author                                          & Time ($\mu$s) & No. iter.\\ \hline
Our $fHf$  & 215 & 155   \\
Our $frHfr$  & 149 & 223   \\ \hline
Valtonen~Örnhag~\etal{}~\cite{valtonenoernhag-etal-arxiv-2020}  & 80 & 416   \\
Ding~\etal{}~\cite{ding-etal-2019-iccv}         & 3301 & 10 \\
Kukelova~\etal{}~\cite{kukelova-etal-2017-cvpr} & 371 & 89  \\ \hline
Fitzgibbon~\etal{}~\cite{fitzgibbon2001}         & 428 & 77 \\
Kukelova~\etal{}~\cite{kukelova2015} & 226 & 147  \\ \hline
\end{tabular}\\
\label{tab:exectime}
\end{table}

\subsection{Real data}\label{sec:realdata}
In this section, we compare the proposed methods on real data. We use the datasets
from~\cite{valtonenoernhag-etal-arxiv-2020}, captured using a UAV with a monochrome
global shutter camera (OV9281) with resolution~$640 \times 480$. The UAV is equipped with an
inertial measurement unit (MPU-9250). In the experiments with only
unknown focal length, the
extracted features where undistorted using a pre-calibrated distortion profile (using the
OpenCV~\cite{opencv} camera calibration procedure); for the case with unknown radial distortion
profile, the raw unprocessed coordinates were used as input.

The ground truth was obtained using a complete SLAM system where the reprojection error and IMU error
were minimized. No scene requirements are enforced by the system, hence feature
points from non-planar structures will be present---such feature points should be discarded by
a robust framework as outliers.

The dataset consists of both indoor and outdoor sequences containing planar surfaces, and
includes varying motions and length of sequences. Example images from the sequences are
shown in~\cref{fig:real_img}.

\def\www{0.1475\textwidth}
\begin{figure}[b]
\centering
\begin{tabular}{cc}
\includegraphics[width=\www]{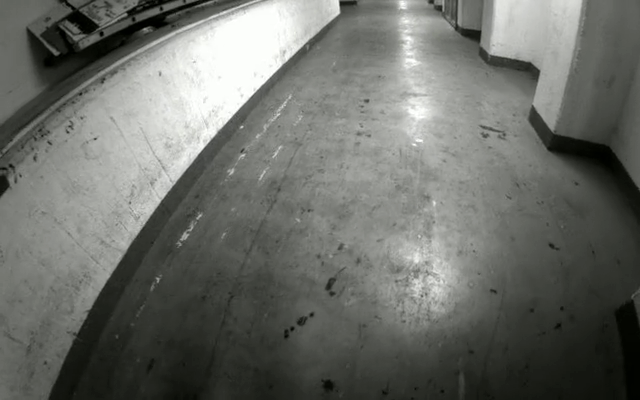} &
\includegraphics[width=\www]{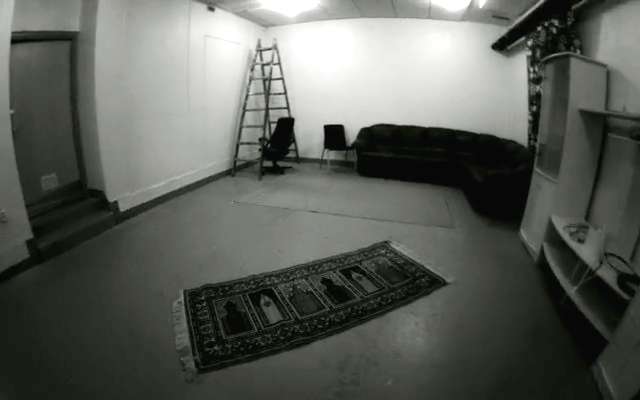} \\\vspace{1mm}
\emph{Basement} &
\emph{Carpet} \\
\includegraphics[width=\www]{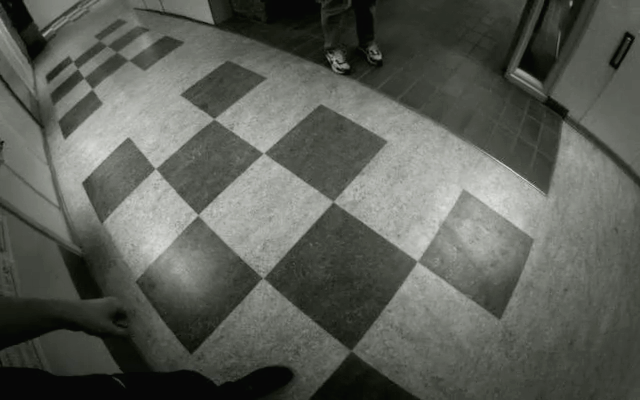} &
\includegraphics[width=\www]{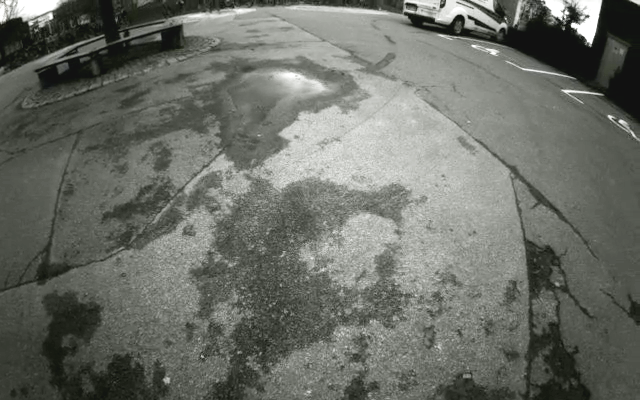} \\
\emph{Indoor} &
\emph{Outdoor} \vspace{-2mm}
\end{tabular}
\caption{Example images from the dataset~\cite{valtonenoernhag-etal-arxiv-2020}.}
\label{fig:real_img}
\end{figure}

We use the IMU filter technique~\cite{5975346}, described in~\cref{sec:embracing}, to obtain the
estimates. Since these measurements are noisy, we propose to use the novel solvers in a
LO-RANSAC framework~\cite{chum-etal-2003}. As a
first step the solvers are used to discard outliers and in the inner LO loop we propose to
optimize over the space of Euclidean homographies with unknown focal length. This refinement step
allows for correction of the errors initially caused by the IMU filter.
Empirically, we have seen that this improves the accuracy.

For a fair comparison, we simulate a scenario where the UAV uses a frame rate of 30 fps,
and limit the number of RANSAC cycles to fit this time frame. For simplicity, we use the values
from~\cref{tab:exectime} alone, acknowledging there
are other parts in the pipeline---such as image capturing, feature extraction and matching,
LO cycles, etc.---that would affect a complete system; however, we argue that this overhead time
is roughly independent of the solver used.

When working with radial distortion correction, one may choose to minimize the reprojection
error in the undistorted image space, or in the distorted image space.
In~\cite{kukelova-larsson-cvpr-2019}, it was shown that it is beneficial to perform
triangulation in the distorted image space. Therefore, we chose to measure the reprojection
in the distorted space.

\subsubsection{Image stitching}
As argued in~\cref{sec:argument} we cannot decompose the homographies obtained
from~\cite{kukelova2015,fitzgibbon2001}, into a relative pose; however, we can still
test the ability of the methods to return an accurate distortion profile.

We simulate a scenario where a UAV is navigating in 30 fps, and limit the number of iterations
for each method according to~\cref{tab:exectime}.
We use the same pixel threshold for all methods, for two consecutive
keyframes of the~\emph{Indoor} sequence.
We chose this sequence, because it naturally contains a checkerboard pattern, which
facilitates in making an accurate ocular evaluation of the quality of the estimated distortion
profile. Physically straight lines should be mapped to straight lines if the rectification is
successful.

\begin{figure*}[t]
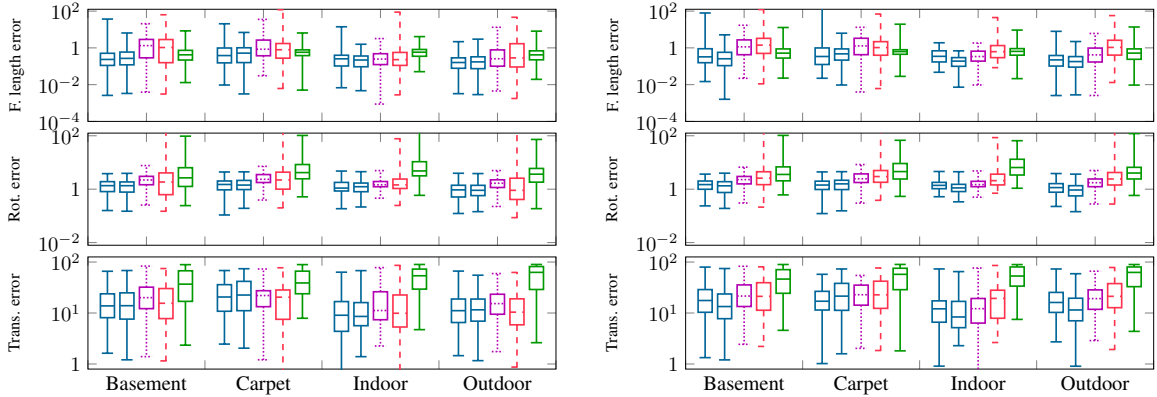

\centering
\def\ytickfontsize{\normalsize}
\def\ylabelfontsize{\footnotesize}

\pgfplotsset{
    boxplot/draw direction=y,
    label style={font=\ylabelfontsize},
    height=\boxplotheightreal,
    xmin=0,xmax=4,
    cycle list={{ourcolor, \thicknesslineplot},%
                {ourcolor, \thicknesslineplot},%
                {valtonenoernhagcolor, densely dotted, \thicknesslineplot},%
                {dingcolor, dashed, \thicknesslineplot},%
                {kukelovacolor, \thicknesslineplot}%
    },
    boxplot/every box/.style={solid},
    boxplot prepared={
        draw position={1/6 + floor(\plotnumofactualtype/5) + 1/6*fpumod(\plotnumofactualtype,5)},
        box extend=0.12,
    },
    x=2cm,
    xticklabels={}, 
    y tick label style={
        font=\ytickfontsize,
    },
}

\begin{tikzpicture}

\newcommand{\realboxplot}[1]{
\input{../graphs/real_data/real_data_hist_map_2020_09_21_21_42_03_#1.txt}   
\input{../graphs/real_data/real_data_hist_map_2020_09_21_21_25_55_#1.txt}   
\input{../graphs/real_data/real_data_hist_map_2020_09_21_20_52_11_#1.txt}   
\input{../graphs/real_data/real_data_hist_map_2020_09_21_22_05_41_#1.txt}   
}

\begin{axis}[
ylabel={F. length error},
ymin=-4,ymax=2.1,
ytick={-4,-2,0,2},
yticklabels={%
        {$10^{-4}$},%
        {$10^{-2}$},%
        {$1$},%
        {$10^{2}$},%
},
name=focal,
]

\input{../graphs/real_data/real_data_hist_map_2020_09_21_21_42_03_f_dist_all.txt}   
\input{../graphs/real_data/real_data_hist_map_2020_09_21_21_25_55_f_dist_all.txt}   
\input{../graphs/real_data/real_data_hist_map_2020_09_21_20_52_11_f_dist_all.txt}   
\input{../graphs/real_data/real_data_hist_map_2020_09_21_22_05_41_f_dist_all.txt}   

\end{axis}
\begin{axis}[
at=(focal.south),
anchor=north,
name=rot,
yshift=-2mm,
ylabel={Rot. error},
ymin=-2.1,ymax=2.1,
ytick={-2,0,2},
yticklabels={%
        {$10^{-2}$},%
        {$1$},%
        {$10^{2}$},%
},
ylabel shift=0.7mm,
]

\input{../graphs/real_data/real_data_hist_map_2020_09_21_21_42_03_R_dist_all.txt}   
\input{../graphs/real_data/real_data_hist_map_2020_09_21_21_25_55_R_dist_all.txt}   
\input{../graphs/real_data/real_data_hist_map_2020_09_21_20_52_11_R_dist_all.txt}   
\input{../graphs/real_data/real_data_hist_map_2020_09_21_22_05_41_R_dist_all.txt}   

\end{axis}
\begin{axis}[
at=(rot.south),
anchor=north,
yshift=-2mm,
ymin=-0.1,ymax=2.1,
ylabel={Trans. error},
xtick={0,1,2,3,4},
x tick label as interval,
xticklabels={%
        {Basement},%
        {Carpet},%
        {Indoor},%
        {Outdoor},%
},
x tick label style={
        text width=2.5cm,
        align=center,
        font=\normalsize
},
ytick={0,1,2},
yticklabels={%
        {$1$},%
        {$10^{1}$},%
        {$10^{2}$},%
},
ylabel shift=2.9mm,
]

\input{../graphs/real_data/real_data_hist_map_2020_09_21_21_42_03_t_dist_all.txt}   
\input{../graphs/real_data/real_data_hist_map_2020_09_21_21_25_55_t_dist_all.txt}   
\input{../graphs/real_data/real_data_hist_map_2020_09_21_20_52_11_t_dist_all.txt}   
\input{../graphs/real_data/real_data_hist_map_2020_09_21_22_05_41_t_dist_all.txt}   

\end{axis}

\pgfplotsset{}

\end{tikzpicture}
\def\ytickfontsize{\normalsize}
\def\ylabelfontsize{\footnotesize}

\pgfplotsset{
    boxplot/draw direction=y,
    label style={font=\ylabelfontsize},
    height=\boxplotheightreal,
    xmin=0,xmax=4,
    cycle list={{ourcolor, \thicknesslineplot},%
                {ourcolor, \thicknesslineplot},%
                {valtonenoernhagcolor, densely dotted, \thicknesslineplot},%
                {dingcolor, dashed, \thicknesslineplot},%
                {kukelovacolor, \thicknesslineplot}%
    },
    boxplot/every box/.style={solid},
    boxplot prepared={
        draw position={1/6 + floor(\plotnumofactualtype/5) + 1/6*fpumod(\plotnumofactualtype,5)},
        box extend=0.12,
    },
    x=2cm,
    xticklabels={}, 
    y tick label style={
        font=\ytickfontsize,
    },
}

\begin{tikzpicture}

\newcommand{\realboxplot}[1]{
\input{../graphs/real_data/real_data_hist_map_2020_09_21_21_42_03_#1_dist_all.txt}   
\input{../graphs/real_data/real_data_hist_map_2020_09_21_21_25_55_#1_dist_all.txt}   
\input{../graphs/real_data/real_data_hist_map_2020_09_21_20_52_11_#1_dist_all.txt}   
\input{../graphs/real_data/real_data_hist_map_2020_09_21_22_05_41_#1_dist_all.txt}   
}

\begin{axis}[
ylabel={F. length error},
ymin=-4,ymax=2.1,
ytick={-4,-2,0,2},
yticklabels={%
        {$10^{-4}$},%
        {$10^{-2}$},%
        {$1$},%
        {$10^{2}$},%
},
name=focal,
]

\input{../graphs/real_data/real_data_hist_map_2020_09_21_21_42_03_f_dist_all.txt}   
\input{../graphs/real_data/real_data_hist_map_2020_09_21_21_25_55_f_dist_all.txt}   
\input{../graphs/real_data/real_data_hist_map_2020_09_21_20_52_11_f_dist_all.txt}   
\input{../graphs/real_data/real_data_hist_map_2020_09_21_22_05_41_f_dist_all.txt}   

\end{axis}
\begin{axis}[
at=(focal.south),
anchor=north,
name=rot,
yshift=-2mm,
ylabel={Rot. error},
ymin=-2.1,ymax=2.1,
ytick={-2,0,2},
yticklabels={%
        {$10^{-2}$},%
        {$1$},%
        {$10^{2}$},%
},
ylabel shift=0.7mm,
]

\input{../graphs/real_data/real_data_hist_map_2020_09_21_21_42_03_R_dist_all.txt}   
\input{../graphs/real_data/real_data_hist_map_2020_09_21_21_25_55_R_dist_all.txt}   
\input{../graphs/real_data/real_data_hist_map_2020_09_21_20_52_11_R_dist_all.txt}   
\input{../graphs/real_data/real_data_hist_map_2020_09_21_22_05_41_R_dist_all.txt}   

\end{axis}
\begin{axis}[
at=(rot.south),
anchor=north,
yshift=-2mm,
ymin=-0.1,ymax=2.1,
ylabel={Trans. error},
xtick={0,1,2,3,4},
x tick label as interval,
xticklabels={%
        {Basement},%
        {Carpet},%
        {Indoor},%
        {Outdoor},%
},
x tick label style={
        text width=2.5cm,
        align=center,
        font=\normalsize
},
ytick={0,1,2},
yticklabels={%
        {$1$},%
        {$10^{1}$},%
        {$10^{2}$},%
},
ylabel shift=2.9mm,
]

\input{../graphs/real_data/real_data_hist_map_2020_09_21_21_42_03_t_dist_all.txt}   
\input{../graphs/real_data/real_data_hist_map_2020_09_21_21_25_55_t_dist_all.txt}   
\input{../graphs/real_data/real_data_hist_map_2020_09_21_20_52_11_t_dist_all.txt}   
\input{../graphs/real_data/real_data_hist_map_2020_09_21_22_05_41_t_dist_all.txt}   

\end{axis}

\pgfplotsset{}

\end{tikzpicture}
\caption{
Errors for the different methods---from left to right: $fHf$, $frHfr$,
Valtonen Örnhag~\etal{}~\cite{valtonenoernhag-etal-arxiv-2020}, Ding~\etal{}~\cite{ding-etal-2019-iccv},
Kukelova~\etal{}~\cite{kukelova-etal-2017-cvpr}---using the metrics~\eqref{eq:errors}.
(Left) rectified input images were used for all but the $frHfr$. (Right) unrectified images
were used for all methods.}
\label{fig:real_boxes}
\end{figure*}

In~\cref{fig:ransac_visual} we show the results of the estimated distortion profile.
We notice that the distortion profile is correct for the proposed method as the yellow area is
a quadrilateral, whereas this is not the case for other methods. Furthermore, we note that
the method by Kukelova~\etal{}~\cite{kukelova2015} does not contain all inliers of the ground plane,
and that the method by Fitzgibbon~\cite{fitzgibbon2001} pick incorrect matches of the wall.

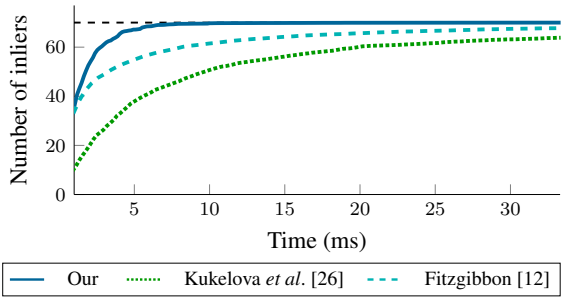
\begin{figure}[h]
\centering
\begin{tikzpicture}[>=latex]
    \begin{axis}[
        width=0.5\textwidth,
        height=0.25\textwidth,
        axis x line*=bottom,
        axis y line*=left,
        ylabel={Number of inliers} ,
        ymin = 0,
        xmin = 1,
        xmax = 33.3,
        xlabel=Time (ms),
        label style={font=\normalsize},
        tick label style={font=\footnotesize},
        y label style={at={(-0.08,0.5)}},
    ]
    \addplot[\mythicknesssmall,color=smallcolor2, dashed] coordinates {
(0.000000,70)
(33.33000,70)
};

    \addplot[\mythickness,color=ourcolor] table {../graphs/real_data/ransac_idx006_time_our.txt};
    \addplot[\mythickness,color=kukelovacolor, densely dotted] table {../graphs/real_data/ransac_idx006_time_kukelova.txt};
    \addplot[\mythickness,color=fitzgibboncolor, dashed] table {../graphs/real_data/ransac_idx006_time_fitzgibbon.txt};
\end{axis}
\end{tikzpicture}
\includestandalone[width=0.425\textwidth]{./graphs/ransac_plot/ransac_legend}
\caption{Number of inliers over time (average of 100 tests). The proposed method is the only method to converge to the correct number of inliers (black dashed line) within the 33.3~ms allotted per frame when running at 30 fps.}
\label{fig:ransac}
\end{figure}

For the same pair of images we measure the inliers as a function of time, see~\cref{fig:ransac}.
The only method converging to the correct number of inliers in the allotted time is our method,
which it does by a large margin.

\subsubsection{Pose estimation}
In this final section, we compare both proposed methods
to~\cite{kukelova-etal-2017-cvpr,ding-etal-2019-iccv,valtonenoernhag-etal-arxiv-2020}. Note
that only one of these methods (our~$frHfr$) estimates the radial distortion profile. As argued
in~\cref{sec:argument}, the methods~\cite{kukelova2015,fitzgibbon2001} cannot estimate the motion
parameters without additional requirements that are not applicable for UAVs, hence cannot be compared in this section.

The error metrics are defined as in~\cite{saurer-etal-2017,ding-etal-2019-iccv,valtonenoernhag-etal-arxiv-2020}, and are given by
\begin{equation}\label{eq:errors}
\begin{aligned}
    e_{\mat{R}} &= \arccos\!\left(\fr{\tr(\mat{R}_{\mathrm{GT}} \mat{R}_{\mathrm{est}}^\T) - 1}{2}\right), \\
    e_{\mat{t}} &= \arccos\!\left(\fr{\mat{t}_{\mathrm{GT}}^\T\mat{t}_{\mathrm{est}} }{\norm{\mat{t}_{\mathrm{GT}}} \norm{\mat{t}_{\mathrm{est}}}}\right), \\
    e_{f} &= \fr{|f_{\mathrm{GT}} - f_{\mathrm{est}}|}{f_{\mathrm{GT}}}\;.
\end{aligned}
\end{equation}

In~\cref{fig:real_traj} we compare the estimated trajectories for all methods. It can be seen
that there are only small differences between the methods using pre-calibrated radial distortion
profile, and the proposed $frHfr$ solver. Furthermore, we measure the errors, according
to~\eqref{eq:errors} for all four sequences. In the left part of~\cref{fig:real_boxes} we
use rectified images for all except the $frHfr$ method, which still performs best or on par
with the other methods in terms of all errors. In the right part of the figure, we run the same
experiment, but all methods are given the raw (unrectified) images as input---here it is clear
that out method achieves superior results.

\section{Conclusions}
We have presented the first ever method capable of simultaneously
estimating the distortion profile, focal length and motion parameters
from a pair of homographies, while incorporating the IMU data.
The method relies on a novel assumption that the IMU data is accurate enough,
to disregard the IMU drift for small time frames, allowing for simpler equations
and faster solvers. We have shown that this assumption is true on both
synthetic and real data, and that the proposed methods are robust.
The method has been shown to give accurate reconstructions, and performs
on par or better than state-of-the-art methods relying on pre-calibration
procedures, while being fast enough for real-time applications.

\clearpage
{\small
\bibliographystyle{ieee_fullname}
\bibliography{uav}
}

\end{document}